# Classification of Polarimetric SAR Images Using Compact Convolutional Neural Networks

Mete Ahishali[a]*, Serkan Kiranyaz[b], Turker Ince[c], Moncef Gabbouj[a]

[a]*Computing Sciences, Tampere University, Tampere, Finland;* [b]*Electrical Engineering Department, Qatar University, Doha, Qatar;* [c]*Electrical and Electronics Engineering Department, Izmir University of Economics, Izmir, Turkey*

*Corresponding author. Email: mete.ahishali@tuni.fi

# Classification of Polarimetric SAR Images Using Compact Convolutional Neural Networks

Classification of polarimetric synthetic aperture radar (PolSAR) images is an active research area with a major role in environmental applications. The traditional Machine Learning (ML) methods proposed in this domain generally focus on utilizing highly discriminative features to improve the classification performance, but this task is complicated by the well-known "curse of dimensionality" phenomena. Other approaches based on deep Convolutional Neural Networks (CNNs) have certain limitations and drawbacks, such as high computational complexity, an unfeasibly large training set with ground-truth labels, and special hardware requirements. In this work, to address the limitations of traditional ML and deep CNN based methods, a novel and systematic classification framework is proposed for the classification of PolSAR images, based on a compact and adaptive implementation of CNNs using a sliding-window classification approach. The proposed approach has three advantages. First, there is no requirement for an extensive feature extraction process. Second, it is computationally efficient due to utilized compact configurations. In particular, the proposed compact and adaptive CNN model is designed to achieve the maximum classification accuracy with minimum training and computational complexity. This is of considerable importance considering the high costs involved in labelling in PolSAR classification. Finally, the proposed approach can perform classification using smaller window sizes than deep CNNs. Experimental evaluations have been performed over the most commonly-used four benchmark PolSAR images: AIRSAR L-Band and RADARSAT-2 C-Band data of San Francisco Bay and Flevoland areas. Accordingly, the best obtained overall accuracies range between 92.33 - 99.39% for these benchmark study sites.

**Introduction**

Synthetic Aperture Radar (SAR) imaging is a widely used remote sensing method in obtaining physical information from the surface regardless of weather conditions. Polarimetric SAR (PolSAR) helps to extract further information using multiple orthogonal polarizations. The classification of PolSAR images is an interesting and motivating task with significance for ecological and socioeconomic applications, and various methods have been proposed such as (Uhlmann and Kiranyaz 2014; Amelard, Wong, and Clausi 2013; Kiranyaz et al. 2012; Yu, Qin, and Clausi 2012; Ince, Ahishali, and Kiranyaz 2017; Ahishali, Ince, et al. 2019). In general, the performance of such traditional methods usually depends on the manually selected features used for the classification. Nevertheless, the method in (Uhlmann and Kiranyaz 2014) has achieved promising classification performances for PolSAR data. Accordingly, they propose to use a large ensemble of classifiers over many electromagnetic and image processing features in high dimensions (e.g., > 100-D). This approach reduced the performance degradation due to the selection of certain features, but the usage of many classifiers and features has significantly increased the computational complexity. Moreover, the classification accuracy of certain class types may still suffer from the suboptimal performance of such fixed and manually selected features.

To address the aforementioned limitation of the traditional methods, in recent years, proposed methods based on (deep) Convolutional Neural Networks (CNNs) have become the *de-facto* standard for many visual recognition applications (e.g., object recognition, segmentation, tracking), achieving the highest performance levels (Lecun, Bengio, and Hinton 2015; Sandler et al. 2018; Tan and Le 2019). However, the deep learners require data of massive sizes, e.g., in the "Big Data" scale to achieve such performance levels. This is also the case in many recent SAR data classification studies based on deep learning (Zhou et al. 2019; Sonobe et al. 2017; Zhou et al. 2016; Gao et al. 2017; S. H. Wang et al. 2018; Lin et al. 2016). For example, in the study (Zhou et al. 2016), the authors propose a classification system that uses 78-80% of the SAR data for training. Other proposed classification methods in (Gao et al. 2017; S. H. Wang et al.

2018) use 75% and the method in (Lin et al. 2016) uses 88 – 92% of all available data in training. Overall, it was necessary to use a significant proportion (e.g., 75% or even higher) of the SAR data to train the deep CNN, but and in practice, such training may not be feasible. Moreover, the output of the deep classifier may no longer be useful or basically needed after providing manual labels for such a large portion of the SAR data. Finally, deep CNNs require a special hardware setup for training and classification to cope up with the high computational complexity required for the deep network structure, which may prevent their use in low-cost or real-time applications.

To address the drawbacks and limitations of the traditional and deep learning based approaches, in this paper, a systematic approach is proposed for accurate classification of the PolSAR data using a compact and adaptive CNN. The proposed method can work directly over the second-order descriptors of PolSAR data without the need for feature extraction and data pre-processing, since CNNs can fuse and simultaneously optimize the feature extractions and classification in a single learning body. Such compact and adaptive CNN is used in the prior work (Ahishali, Kiranyaz, et al. 2019) for dual- and single-polarized SAR image classification. In this study, the proposed approach is used for the PolSAR classification task. In this way, the competing methods can now utilize a large number of additional features derived from different Target Decomposition theorems, due to the increased number of polarizations.

In this study, it will be shown that the proposed compact and adaptive CNNs can achieve improved PolSAR classification performance levels with an insignificant amount of training data (e.g., <0.1% of the entire SAR data) and superior computational complexity appropriate for real-time processing. Moreover, the usage of very low-resolution (e.g., $7 \times 7$ to $19 \times 19$ pixel sliding window) patches no longer poses a problem due to the compact nature of the proposed CNN configuration.

**SAR Data Processing**

PolSAR systems measure backscattering $[S]$ matrix, which is complex and produced by the observed target. If the assumption is made about having linear polarizations horizontally and vertically for transmitting and receiving, $[S]$ can be expressed as

$$E^r = [S]\begin{bmatrix} E_h^t \\ E_v^t \end{bmatrix}, where\ [S] = \begin{bmatrix} S_{hh} & S_{hv} \\ S_{vh} & S_{vv} \end{bmatrix}$$

$$\Rightarrow E^r = \begin{bmatrix} |S_{hh}|e^{j\phi_{hh}} & |S_{hv}|e^{j\phi_{hv}} \\ |S_{vh}|e^{j\phi_{vh}} & |S_{vv}|e^{j\phi_{vv}} \end{bmatrix}\begin{bmatrix} E_h^t \\ E_v^t \end{bmatrix}, \quad (1)$$

where $S_{hv} = S_{vh}$ holds for monostatic system configurations using reciprocity theorem and $E^r$ and $E^t$ are the received and transmitted electric fields. As the main consequence, a given target is now represented by five parameters (Lee and Pottier 2009): the three absolutes ($|S_{hh}|$, $|S_{vv}|$, and $|S_{hv}|$ or $|S_{vh}|$) and the two relative phases ($\phi_{hv-hh}$, $\phi_{vv-hh}$).

*PolSAR Information Extraction*

Second-order polarimetric representations can be used to have a better extraction of physical information from the $2 \times 2$ backscattering $[S]$ matrix. Due to speckle noise and random scattering, PolSAR data are generally acquired by multilooking, where the processed data are obtained by averaging *n*-looks. Accordingly, the average polarimetric covariance $\langle[C]\rangle$ and coherency $\langle[T]\rangle$ matrices can be written using Lexicographic scattering and Pauli based scattering vectors, respectively. Multi-look coherency matrix $\langle[T]\rangle$ can be represented as

$$\langle[T]\rangle = \frac{1}{n}\sum_{i=1}^{n} k_i k_i^{*T}, \quad (2)$$

where $k_i$ is Pauili based scattering vector for each look, $k_i = [S_{hh} + S_{vv}, S_{hh} - S_{vv}, 2S_{hv}]^T/\sqrt{2}$. Using the lexicographic basis, $\Omega$, the covariance matrix is obtained as $\langle[C]\rangle = \frac{1}{n}\sum_{i=1}^{n}\Omega_i\Omega_i^{*T}$ where $\Omega = [S_{hh}, \sqrt{2}S_{hv}, S_{vv}]^T$. It can be said that both coherency $\langle[T]\rangle$ and covariance $\langle[C]\rangle$ matrices are equivalent in terms of polarimetric description information of the target since they are both $3 \times 3$ Hermitian positive definite matrices and they can be linearly transformable from one to another.

Another target descriptor common in SAR image processing is the total scattering power (SPAN) information. Hence, the Frobenius norm (SPAN) of the scattering matrix $\langle[S]\rangle$ from Eq. (1) can be expressed as (Lee and Pottier 2009);

$$Span(S) = Tr(S\,S^{*T})$$
$$= |S_{hh}|^2 + |S_{vh}|^2 + |S_{hv}|^2 + |S_{vv}|^2 \quad (3)$$

Additionally, in PolSAR classification, classifiers can utilize various target decompositions (TDs). TDs consist of coherent and incoherent types of decompositions. Coherent TDs aim to represent backscattering coefficients as the sum of independent components, such as the Pauli decomposition, Krogager decomposition (Krogager 1990), and Cameron decomposition (Cameron and Leung 1990). Basically, backscattering coefficients, $\langle[S]\rangle$ in Eq. (1), has complete information of the observed target, where different interpretations of $\langle[S]\rangle$ may yield having more discriminative representations. Hence, these independent components are used in SAR applications to have more representative power of the target. On the other hand, incoherent TDs exploit distributed scatters based on the incoherently averaged second-order descriptors, $\langle[C]\rangle$ and $\langle[T]\rangle$ matrices. For example, some of incoherent TDs are Huynen decomposition (Huynen 1970), Cloude-Pottier (eigenvector-eigenvalue or H/$\alpha$/A) decomposition (Cloude and Pottier 1997), and Freeman decomposition (Freeman and Durden 1998).

*Prior Work*

Many traditional approaches for classification of SAR data have been applied (Uhlmann and Kiranyaz 2014; L. Chen et al. 2010; Yang, Zou, et al. 2010; Gigli, Sabry, and Lampropoulos 2007)**.** Most utilized various PolSAR features based on combinations of different TDs. There are two categories of SAR features used by the most promising traditional classifiers, such as Support Vector Machines (SVMs) (Shimoni et al. 2009; L. Chen et al. 2010) and Random Forest (RF) (Gigli, Sabry, and Lampropoulos 2007; Yang, Zou, et al. 2010). The first category consists of features extracted directly from SAR data or its scattering matrix and the second-order descriptors, $\langle[C]\rangle$ and $\langle[T]\rangle$. The second consists of the methods that use features extracted using target decomposition theorems.

In PolSAR classification, Deep Learning paradigms constitute the new trend with deep CNNs. The study in (S. H. Wang et al. 2018) proposes a very deep CNN architecture consisting of 11 layers. They use Pauli intensity channels $k_i = [S_{hh} + S_{vv}, S_{hh} - S_{vv}, 2S_{hv}]^T/\sqrt{2}$ as in Eq. (2) to build a 3-channel CNN input. Another study in (Zhou et al. 2016) proposes to use a 6-dimensional real vector as the input of CNN classifier, formed using both diagonal elements and magnitudes of the non-diagonal complex elements of the coherency matrix $\langle[T]\rangle$. Another recent work in (Gao et al. 2017) proposes a dual-branch deep CNN for the classification task; two CNNs are trained together, but over different inputs: one with 6-channel electromagnetic features as similar to the method in (Zhou et al. 2016), and the other with 3-channel RGB images. However, as previously mentioned, all these deep learning based methods require a large amount of training data to obtain acceptable classification performances, and this may undermine the aim of the automatic classification system. Moreover, the limited test data used in the above-mentioned studies indicate that the deep CNN based methods have not been tested over the majority of the SAR data, and this may further create a reliability issue on the performance level achieved.

The method proposed in (Uhlmann and Kiranyaz 2014) has provided elegant classification performance using a significantly smaller number of training samples (less than 0.1% of the benchmark PolSAR data) and with a minimum computational complexity compared to deep CNNs. Based on Table 1, in this study, the method in (Uhlmann and Kiranyaz 2014) is considered as the leading approach in PolSAR classification when compared to Machine Learning approaches using a smaller number of training samples. Note that for Flevoland, AIRSAR data, the proposed method in (Uhlmann and Kiranyaz 2014) has 15 different class types whereas the others tend to have fewer classes. The method proposed in that study is based on combining multiple image processing features (texture and color features) with electromagnetic features extracted by coherent and incoherent TDs. Then, to maximize classification accuracy, an ensemble of conventional classifiers is used to learn all these features simultaneously.

Table 1. Classification methods proposed in previous studies and their achieved classification accuracies reported for different PolSAR data.

|  | San Francisco, RADARSAT2 | Flevoland, AIRSAR | Flevoland, RADARSAT2 |
|---|---|---|---|
| (Ren et al. 2019) | 0.9076 | - | 0.8987 |
| (W. Chen et al. 2018) | - | 0.9473 | 0.9482 |
| (Ren et al. 2020) | - | 0.8592 | - |
| (L. Wang et al. 2018) | - | - | 0.944 |
| (H. Liu et al. 2016) | - | - | 0.9004 |
| (Uhlmann and Kiranyaz 2014) | 0.9225 | 0.9169 | 0.9568 |

**Methodology**

The proposed systematic approach for classification is illustrated in Figure 1 over the benchmark *SFBay_C* (San Francisco, C band) PolSAR data. In order to perform classification, an $N \times N$ window of each individual electromagnetic (EM) channel around each pixel has been fed to a compact an adaptive 2D CNN as a distinct input. Accordingly, the corresponding output of CNN determines its center pixel's label. Therefore, the number of the utilized EM channels determines the size of the input layer of CNN. In this work, 3 to 6 EM channels have been tested. One hyper-parameter in this model is the size ($N$) of the $N \times N$ sliding window. In deep CNN approaches, a high $N$ has is necessary. However, the proposed compact topology enables using small windows such as $7 \times 7$ or even $5 \times 5$ pixels that can affect the classification performance.

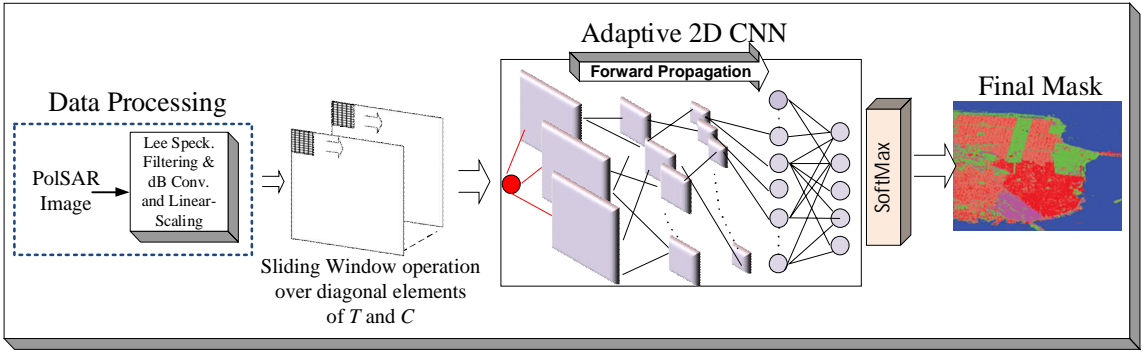

Figure 1. The proposed classification system for PolSAR data based on compact CNNs using sliding-window operation. Linear-scaling is applied after dB conversion to have pre-processed data in the range of $[-1, 1]$ for the CNN.

*Adaptive CNN Implementation*

In the proposed adaptive CNN implementation, there are two types of hidden layers: 1) CNN layers into which conventional "convolutional" and "subsampling-pooling" layers are merged and 2) fully-connected (or MLP) layers. Therefore, neurons of the hidden CNN layers are modified such that each neuron is capable of performing both convolution and down-sampling. The intermediate outputs of each neuron are sub-sampled to obtain the final output of that particular neuron. Then, the final output maps are convolved with their individual kernels and further cumulated to form the input of the next layer neuron. Accordingly, the final output of the $k^\text{th}$ neuron at layer $l$, $s_k^l$, is the sub-sampled version of the intermediate output $y_k^l$. The input map of the next layer neuron will be obtained by the cumulation of the final output maps of the previous layer neurons convolved with their individual kernels as follows:

$$x_k^l = b_k^l + \sum_{i=1}^{N_{l-1}} conv2D(w_{ik}^{l-1}, s_i^{l-1}, 'NoZeroPad') \quad (4)$$

Each input neuron in the input layer is fed with the patch of the particular channel. As discussed earlier, in this study the number of channels is varied from 3 to 6. The adaptivity of the proposed compact CNNs enables the user to select different patch (sliding window) sizes that highly depends on the SAR data and selected experimental setup. Moreover, the number of hidden CNN layers can be set to any number, regardless of the input patch size. This ability is possible in this implementation because the sub-sampling factor of the output CNN layer (the last hidden CNN layer just before the first MLP layer) is set to the dimensions of its input map.

*Back-Propagation for Adaptive CNNs*

The illustration of the Back-Propagation (BP) training of the adaptive CNNs is shown in Figure 2. For an $N_L$-class problem, the class labels are first converted to the target class vectors using 1-of-$N_L$ encoding scheme. Then, for each window (patch) with its corresponding target and output class vectors, $[t_1, \ldots, t_{N_L}]$ and $[y_l^L, \ldots, y_{N_L}^L]$, respectively, the main interest is to find the derivative of this error with respect to each individual network parameter (weights and biases). Let $l=1$ and $l=L$ be the input and output layers, respectively. The error (MSE) in the output (MLP) layer can be expressed as:

$$E = E(y_1^L, \ldots, y_{N_L}^L) = \sum_{i=1}^{N_L}(y_i^L - t_i)^2 \quad (5)$$

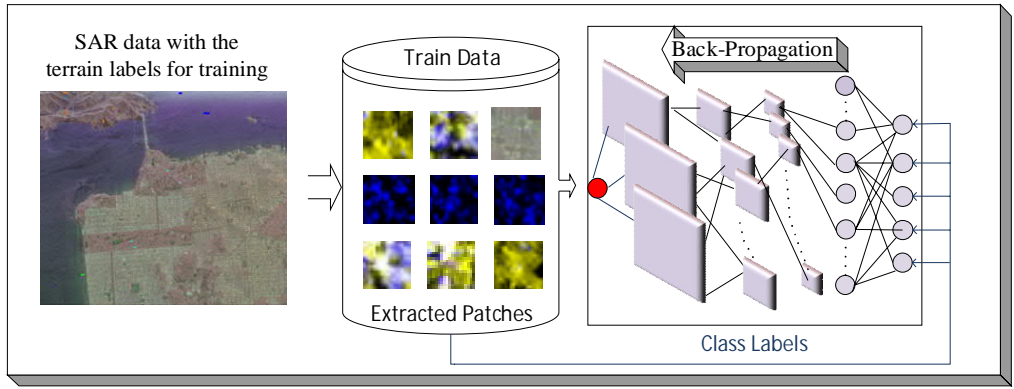

Figure 2. The training process of the adaptive 2D CNN with ground-truth labels over PauliRGB coded SAR image after pre-processing (logarithmic transform and scaling).

The already well-studied BP formulation of the CNN layers is omitted here. In general, the BP training of the CNN layers is composed of 4 distinct operations: inter-BP among CNN layers, intra-BP within a CNN neuron, BP from the first MLP layer to the last CNN layer, and computation of the weight (kernel) and bias sensitivities.

**Experimental Results**

First, the configured experimental setup will be presented along with benchmark PolSAR data used in this study. Next, there will be a discussion of the proposed approach with the compact and adaptive 2D CNNs compared to that of the method in (Uhlmann and Kiranyaz 2014) and other more recent methods. As the visual evaluation is also possible over the segmentation mask resulting from the classification, masks of each study site will be presented. In this way, this study will provide assessment and evaluation in terms of both quantitative and visual (qualitative) performance.

*Benchmark PolSAR Data*

Four benchmark PolSAR data are used for testing and comparative evaluations. Two were acquired by the airborne system, NASA/JET Propulsion Laboratory AIRSAR ("ESA PolSARPro, Sample Datasets" n.d.), and further two, by the spaceborne system, Canadian

Space Agency RADARSAT-2 (Moon et al. 2010). Slant range geometry is used for all study sites in the experimental evaluations. The details of these benchmark SAR data are presented in Table 2.

Table 2. Polarimetric SAR Images used in this work.

| Name | System&Band | Abbr. | Date | Incident angles |
|---|---|---|---|---|
| SF Bay | AIRSAR L | *SFBay_L* | 1988 | 10 - 60° |
| SF Bay | RADARSAT-2 C | *SFBay_C* | Apr 2008 | 30° |
| Flevoland | AIRSAR L | *Flevo_L* | Aug 1989 | 40 - 50° |
| Flevoland | RADARSAT-2 C | *Flevo_C* | Apr 2008 | 30° |

Table 3. The number of classes, training size, and GTD size.

| Name | Dimensions | #class | Train Size per Class | Total GTD Size |
|---|---|---|---|---|
| *SFBay_L* | $900 \times 1024$ | 5 | ~292 | 123459 |
| *SFBay_C* | $1426 \times 1876$ | 5 | 500 | 252500 |
| *Flevo_L* | $750 \times 1024$ | 15 | 120 - 480 | 209979 |
| *Flevo_C* | $1639 \times 2393$ | 4 | 500 | 202000 |

The first study area consists of the San Francisco Bay located in California, USA, and acquired at L-band (*SFBay_L*) & C-band (*SFBay_C*). The second is the Flevoland area in the Netherlands at L-band (*Flevo_L*) & C-band (*Flevo_C*). They are the most commonly used benchmark data in PolSAR classification problems. The Flevoland area mainly consists of vegetation fields, and the San Francisco Bay area, of urban and natural zones. The sizes of the entire ground truth (GTD) and the train data are presented in Table 3 for each study site. It is difficult to provide 100% accurate land cover definitions, therefore, the GTDs used in this work are the same as those in many previous studies to be discussed in the following. Note that if GTD is erroneous, it will equally affect the proposed approach and competing methods, which is a justification for the evaluation of methods with the same GTD.

*San Francisco Bay, AIRSAR, L-Band (SFBay_L)*

This benchmark PolSAR data covers the San Francisco Bay area, providing mainly urban class information combined with natural classes for the experiments. It is fully polarized and has a 900 × 1024 image size, with pixel resolution of 10 × 10 meters. Because there is no available certain ground truth for the data, (Kiranyaz et al. 2012; Uhlmann et al. 2011) constructed a ground truth by visually inspecting PauliRGB coded image of *SFBay_L* based on aerial photographs provided by the TerraServer Web site ("U.S. Geological Survey Images" n.d.). Consequently, the determined 5-classes are water, urban, forest, bare soil, and natural vegetation (scrub and woodland). This ground truth with the specified class types exactly corresponds to the GTD used in (Kiranyaz et al. 2012; Uhlmann et al. 2011; Ince, Ahishali, and Kiranyaz 2017). Similar class definitions are used in many other studies: (X. Liu et al. 2018; Zhang et al. 2019; Huang et al. 2019; Yin, Yang, and Yamaguchi 2009). For *SFBay_L*, selected regions for train and test are shown in Figure 3. To ensure a fair comparison with (Uhlmann and Kiranyaz 2014), in this study, 1% − 2 % of ground truth samples are used for training (~0.1% of the entire data).

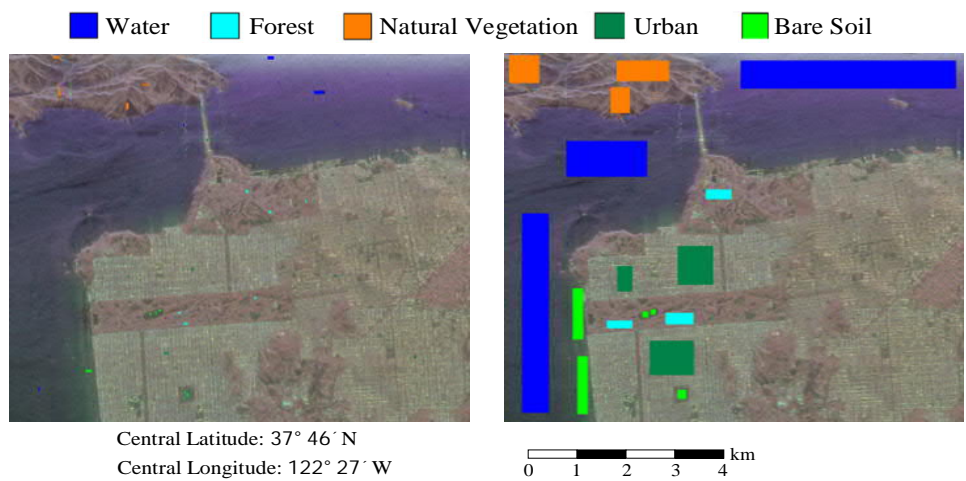

Figure 3. The pixel samples selected for training (left) and the non-overlapping regions for testing (right) over the benchmark *SFBay_L* PolSAR image after pre-processing (logarithmic transform and scaling).

*San Francisco Bay, RADARSAT-2 C-Band (SFBay_C)*

The *SFBay_C* fully polarimetric PolSAR data have similar characteristics with *SFBay_L*, except that the former is spaceborne and has $10 \times 5$ meters resolution. As shown in Figure 4, the same ~$1400 \times 1800$-pixel subregion and land cover definition used here is similar to that presented in (Ren et al. 2019; Y. Chen et al. 2019; X. Liu et al. 2018; Ren et al. 2018; Uhlmann and Kiranyaz 2014). Accordingly, the GTD was constructed by visual inspection of aerial photographs of this area as for *SFBay_L*. There are three major land cover types: water, man-made, and vegetation. The man-made class type is further divided into developed, high-density urban, and low-density urban classes, based on their inclusion of natural classes. Some may argue that there is considerable overlap between low density urban, high density urban, and developed classes. However, visual inspection of Figure 4 shows that there are clear existing differences, and they should therefore be considered as separate classes. High-density and low-density urban classes consist of the pixels from the urban areas; for high-density urban class, the areas are more congested, consisting of more densely concentrated buildings and/or man-made structures compared to the low-density urban class. Developed class samples consist of rather sparse man-made structures mixed with vegetation. Overall, this classification problem can also be considered as a hierarchical classification task, due to the existing hierarchical structure among urban type of classes. Hence, as stated in (Uhlmann and Kiranyaz 2014), the GTD accuracy is not guaranteed, since the different manmade classes may also contain plants and trees (e.g., in the gardens of houses for the developed class type). For *SFBay_C*, 1:100 proportion is followed for the samples used in learning and testing with 500:50K number of samples per class.

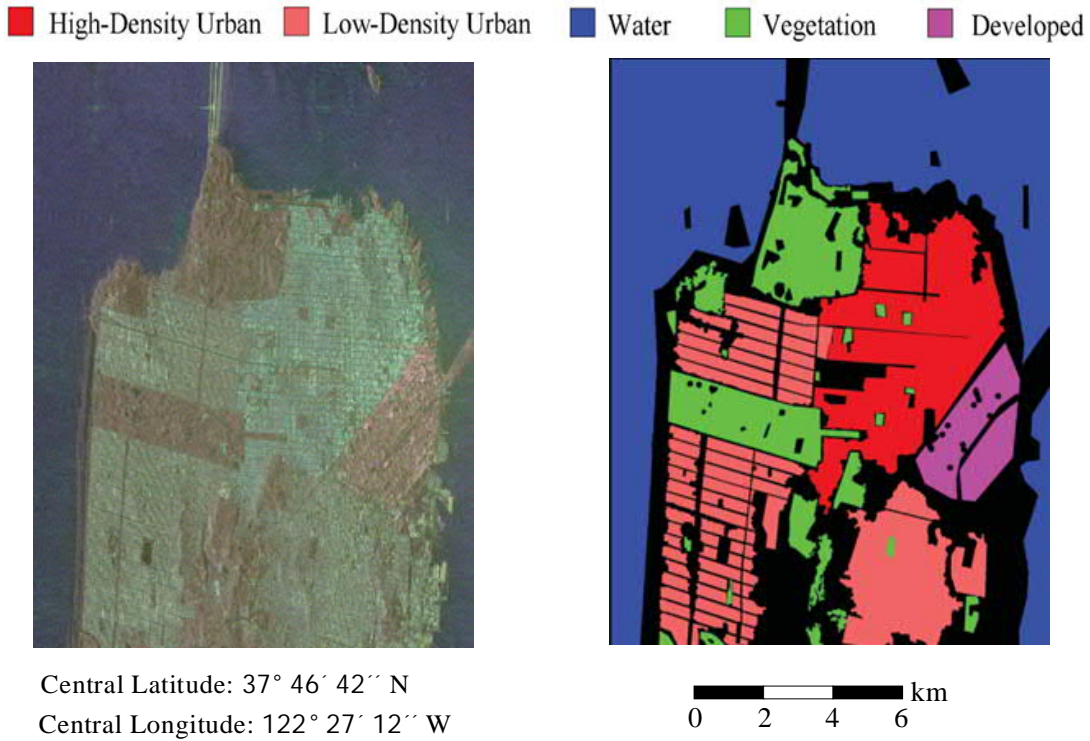

Central Latitude: 37° 46´ 42´´ N
Central Longitude: 122° 27´ 12´´ W

Figure 4. PauliRGB coded *SFBay_C* PolSAR image is given (left) and the corresponding ground truth set (right) is shown with class labels after pre-processing (logarithmic transform and scaling).

*Flevoland, AIRSAR, L-Band (Flevo_L)*

The fully polarimetric *Flevo_L* PolSAR data have $750 \times 1024$ pixels with approximately $12 \times 6$ meters resolution. It was acquired in mid-August 1989 during the MAESTRO-1 Campaign. The classes mainly consist of different vegetation and soil types with a water region, but also, some human-made classes with a few buildings. Thus, it has been used extensively in many crop and land classification applications, and it has well-established ground truth information provided by (Yu, Qin, and Clausi 2012) with 15 classes: water, forest, lucerne, grass, rapeseed, beet, potatoes, peas, stem beans, bare soil, wheat A, wheat B, wheat C, barley, and building. The GTD is shown in Figure 5 by assigning distinct RGB values to each class. In this work, the number of pixels used for the training over *Flevo_L* is varied from 120 to 480 per class, in order to evaluate the effect of training data size over the classification performance.

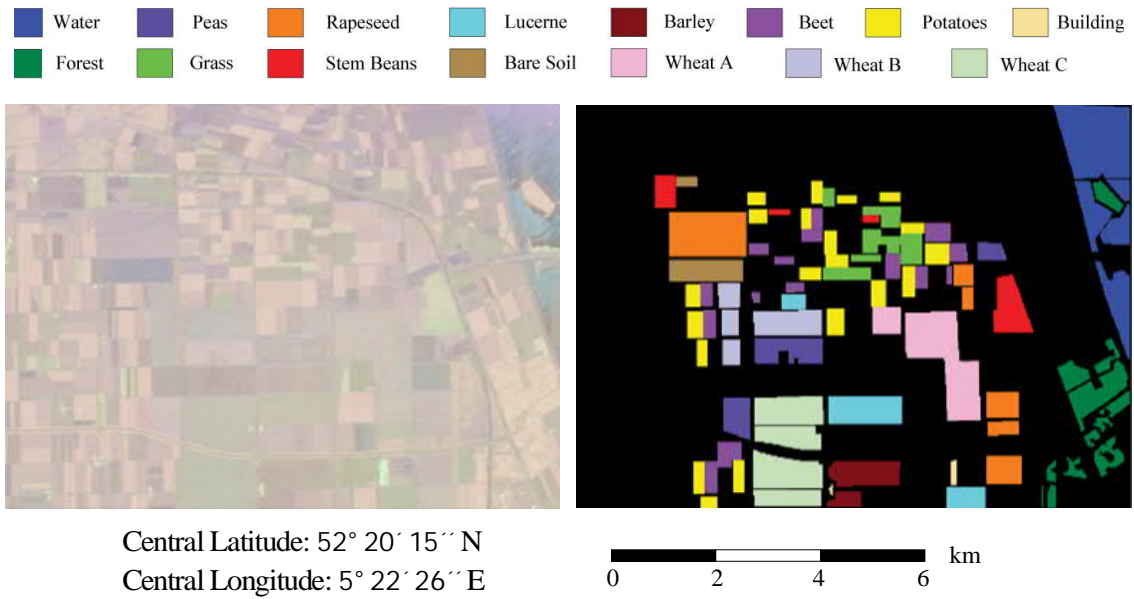

Figure 5. *Flevo_L* PolSAR image after logarithmic transform and scaling (top) and corresponding ground truth land cover (bottom). Each class is assigned to a distinct RGB value for illustration purposes.

*Flevoland, RADARSAT-2, C-Band (Flevo_C)*

Fully polarimetric C-band SAR data having approximately $10 \times 5$ meters resolution of Flevoland, The Netherlands, were acquired in April 2008. This study site has similar characteristics to *Flevo_L*, however with a greater emphasis on human-made class types. As shown in Figure 6, the same ~$1600 \times 2400$-pixel subregion and GTD as presented in (Uhlmann and Kiranyaz 2014) are used consisting of four classes: water, urban, forest, and cropland (Yang, Dai, et al. 2010). Furthermore, the same 1:100 ratio is followed as in (Uhlmann and Kiranyaz 2014) to partition the train and test data.

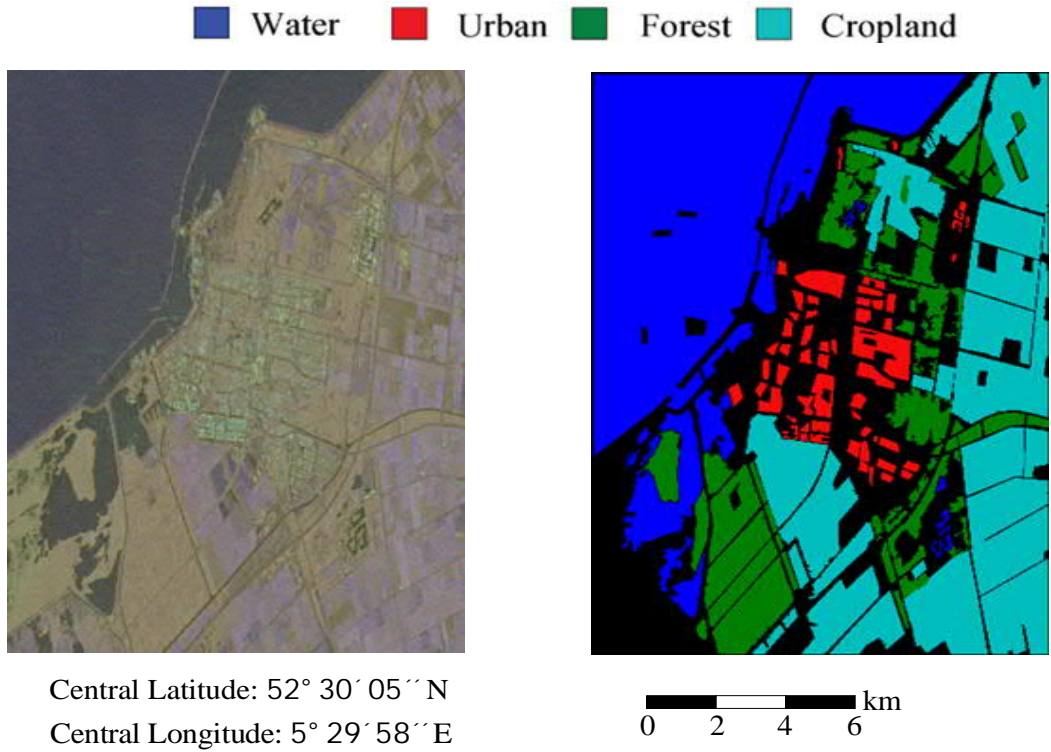

Central Latitude: 52° 30´ 05´´ N
Central Longitude: 5° 29´ 58´´ E

Figure 6. *Flevo_C* PolSAR image after logarithmic transform and scaling (left) and the corresponding ground truth land cover (right) is given. Each class is illustrated with distinct RGB values.

*Experimental Setup*

All four benchmark PolSAR images are originally multi-look averaged by using four look data. Additionally, *Lee's* proposed polarimetric speckle filter (Lee 1999) with a $5 \times 5$ window size is applied to all study sites to provide unbiased comparison with the competing methods in (Kiranyaz et al. 2012) and (Uhlmann and Kiranyaz 2014). Hyper-parameters of the conventional classifiers (SVM and CNBC) and the proposed adaptive CNN are selected by using 50% percent of training data as a validation set. The hyper-parameter search space of SVM for each feature combination in the experiments includes 1) the kernel type: linear, polynomial (3$^{rd}$ order), radial basis function (RBF), and sigmoid, 2) the gamma parameter (for polynomial, RBF, and sigmoid kernels): $1/2^n$ where $n = \{1,2,3,4\}$, and 3) the regularization parameter, C = $1/2^n$ where $n = \{0,1,2,3\}$.

The proposed adaptive 2D CNN is implemented using C++ over MS Visual Studio 2015 in 64 bit. This is a non-GPU implementation; however, Intel ® OpenMP API is used to obtain multiprocessing with shared memory. SVM and CNBC classifiers are also implemented in C++ using MS Visual Studio 2013 in 32 bit. A computer with I7-4790 at 3.6GHz (4 real, 8 logical cores) and 16Gb memory is used in all performed experiments for both training and testing. The proposed CNN configuration has only one convolutional layer, consisting of 20 filters with kernel dimensions, $K_x=K_y=3$, and subsampling factors, $ssx=ssy=2$, and one hidden MLP layer with 10 neurons. The hyperbolic tangent is used as the activation function of all layers.

Due to the compactness of the used CNN configuration, over-fitting does not pose any threat. Therefore, the maximum number of training iterations is used as the sole early stopping criterion, as given in Table 4. Note the fact that, the set maximum number of training iterations is relatively smaller for *SFBay_L* compared to others, since the compact CNN is able to converge within 40 iterations and provide >99% accuracy for *SBay_L*. Learning rate ε is initially set to 0.05 and a dynamic learning rate adaptation is performed in each BP iteration, i.e., if the train MSE decreases in the current iteration, ε is slightly increased by 5%; otherwise, it is reduced by 30% for the next iteration.

Table 4. The maximum number of iterations for each study site as an early stopping criterion.

| Dataset | Maximum Iteration |
|---------|-------------------|
| *SFBay_L* | 40 |
| *SFBay_C* | 400 |
| *Flevo_L* | 600 |
| *Flevo_C* | 400 |

## Results and Performance Evaluations

### Performance Evaluations over SFBay_L

For *SFBay_L*, the comparative evaluations against SVM, CNBC (Kiranyaz et al. 2012), and the method in (Uhlmann and Kiranyaz 2014) are performed by presenting overall classification accuracy and in particular, each individual performance improvement is reported per class. The utilized features in the competing methods are given in Table 5 for the classification. These features are incrementally concatenated to evaluate CNBC and SVM classifier performances. CNBC is built by training a 4-layer MLP (with 16 and 8 hidden neurons) for each binary classifier in the network. Classification accuracies of these competing methods with different feature combinations are presented in Table 6.

Table 5. Target Decompositions (TDs) used in competing methods as features.

|  | **Feature** | **Dimension** |
|---|---|---|
| FV1 | [T] and [C] Matrices | 12 |
| FV2 | Span, H/A/Alpha (Cloude and Pottier 1996) | 7 |
| FV3 | Eigenanalysis - Eigenvalues | 3 |
| FV4 | Correlation Coefficients | 6 |
| FV5 | Touzi (Touzi 2006) | 4 |
| FV6 | Krogager (Krogager 1990) | 3 |
| FV7 | Freeman (Freeman and Durden 1998) | 3 |
| FV8 | Huynen (Huynen 1970) | 3 |
| FV9 | VanZyl (van Zyl 1993) | 3 |
| FV10 | Yamaguchi (Yamaguchi et al. 2005) | 4 |

The classification performance of adaptive CNN is presented in Table 7 for *SFBay_L*. The comparison between Table 6 and Table 7 reveals that the proposed approach with the adaptive 2D CNN outperforms the competing methods despite using only 3-dimensional EM features (diagonal elements of $[T]$), whilst the competing methods use 4 to 12 times more EM features and TDs. In addition to the first setup, the *span* is added to the input layer of the adaptive CNN as the 4[th] channel, and its effect is observed. Moreover, a 6-channel setup is implemented using diagonal elements of both $[T]$ and $[C]$. Accordingly, the proposed approach can achieve an average classification accuracy of 1.32 and 1.37% higher than the competing CNBC and SVM methods, respectively. Note that even though the performance improvements are limited, CNBC and SVM use 48-D and relatively smaller 35-D features, respectively, to achieve such performance levels. However, the proposed approach has at most a 6-channel setup. To achieve a more balanced comparison, when using only FV1 with 12-D features (which is still 3 times the number of features compared to the proposed approach), CNBC and SVM are only able to achieve approximately 96% classification accuracy, which is ~3% less than that obtained by the proposed approach.

Table 6. Classification accuracies of the competing methods (CNBC and SVM) with different feature combinations for *SFBay_L*.

| SFBay (L Band) | Dimension | CNBC | SVM |
|---|---|---|---|
| FV1 | 12 | 0.9583 | 0.9563 |
| FV2+(FV1) | 19 | 0.9723 | 0.9734 |
| FV3+(FV1+FV2) | 22 | 0.9763 | 0.9746 |
| FV4+(FV1+FV2+FV3) | 28 | 0.9786 | 0.9791 |
| FV5+(FV1+…+FV3+FV4) | 32 | 0.979 | 0.9798 |
| **FV6+(FV1+…+FV4+FV5)** | **35** | **0.9759** | **0.9802** |
| FV7+(FV1+…+FV5+FV6) | 38 | 0.9806 | 0.98 |
| FV8+(FV1+…+FV6+FV7) | 41 | 0.9796 | 0.9801 |
| FV9+(FV1+…+FV7+FV8) | 44 | 0.9803 | 0.9798 |
| **FV10+(FV1+...+FV8+FV9)** | **48** | **0.9807** | **0.9792** |

Based on the obtained results for *SFBay_L*, it is observed that the optimal window size varies from *N*=19 to *N*=23, depending on the number of channels. However, if the optimal number of input channels are selected with respect to different window sizes, any window size *N*=9 or higher (except *N*=31, where the correlation within the neighborhood starts decreasing) can yield accuracies above 98.2% which is better than the best accuracy obtained by the competing methods over *SFBay_L*.

Table 7. Classification accuracies of the proposed approach for *SFBay_L* with different window size and channels.

| SFBay (L Band) | 3-channels | 4-channels | 6-channels |
|---|---|---|---|
| **Window Size** | $T_{11}, T_{22}, T_{33}$ | $T_{11}, T_{22}, T_{33}, span$ | $T_{11}, T_{22}, T_{33}, C_{11}, C_{22}, C_{33}$ |
| **7x7** | 0.9748 | 0.9726 | 0.9751 |
| **9x9** | 0.9846 | 0.9825 | 0.9764 |
| **11x11** | 0.9888 | 0.9812 | 0.9843 |
| **13x13** | 0.9807 | 0.9864 | 0.99 |
| **15x15** | 0.9847 | 0.9865 | 0.9918 |
| **17x17** | 0.9888 | 0.9889 | 0.9893 |
| **19x19** | 0.9884 | 0.9915 | **0.9936** |
| **21x21** | 0.9807 | **0.9939** | **0.9934** |
| **23x23** | 0.9911 | 0.9908 | 0.9901 |
| **25x25** | 0.9817 | 0.9777 | 0.9855 |
| **31x31** | 0.97 | 0.963 | 0.9515 |

For visual evaluation, the training samples and the final segmentation masks for *SFBay_L* are shown in Figure 7. An important observation is that the setup with 4-channels and $21 \times 21$ pixels window has the highest accuracy, but the segmentation mask suffers from the coarse visual resolution, whereas the setup with $7 \times 7$ pixels window can achieve finer details.

The confusion matrix produced by the proposed approach is given in Table 8 for *SFBay_L*. For a detailed comparison, the classification accuracies per class type are presented in Figure 8 as well. While the classification performance is similar for some classes (e.g., Water and Urban), significant performance gaps occur, e.g., by > 20% for Forest. This outcome is expected because the competing methods use fixed and manually selected features which are unable to exhibit the same level of discrimination for this class type as for others, unlike the adaptive CNN, which is able to "learn to extract" such features. This is a major advantage in terms of reliability, since it is apparent that the competing methods may fail to classify certain class types with reasonable accuracy.

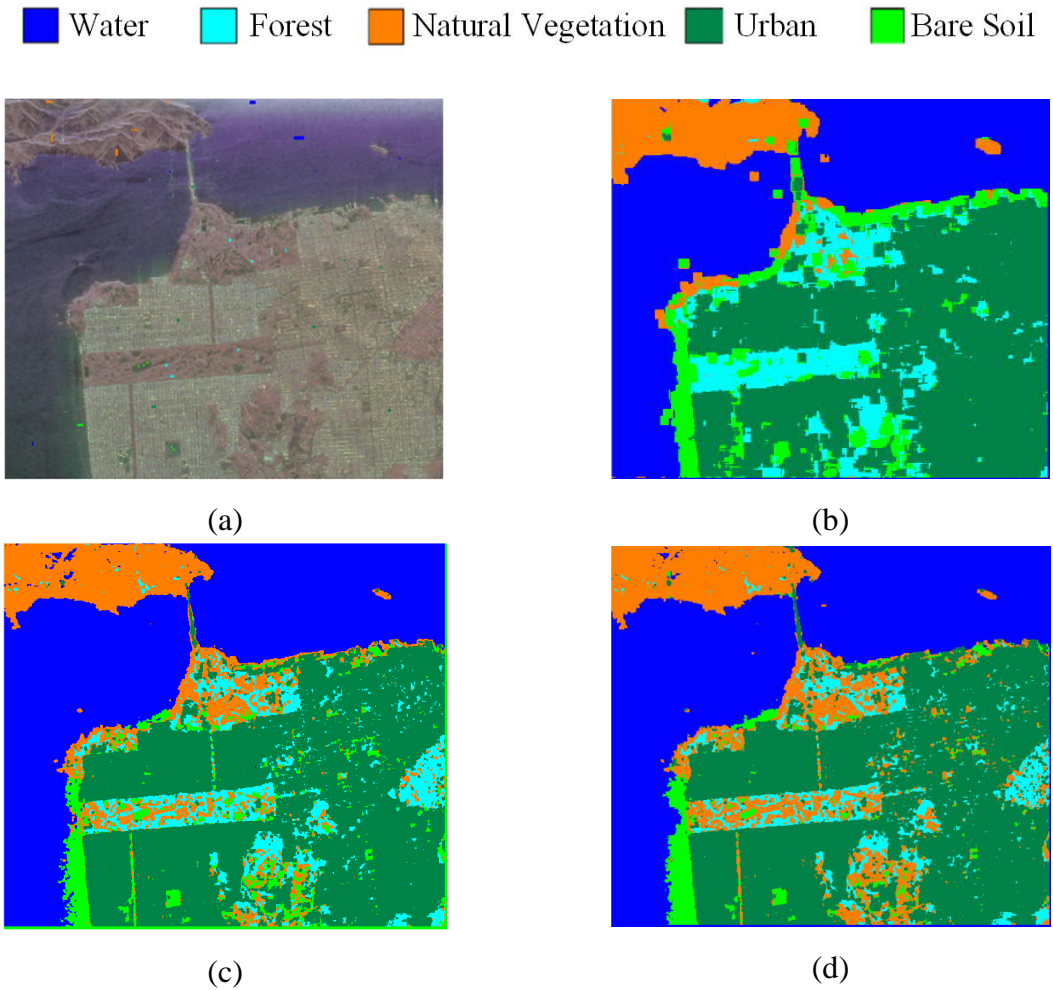

Figure 7. Final segmentation masks by the proposed approach for *SFBay_L* with (a) pixels in the train data. The segmentation masks obtained using 4-channel input with $21 \times 21$ and $7 \times 7$ pixels windows are shown in (b) and (c), respectively. In (d), the segmentation mask by 3-channel input with $7 \times 7$ pixels window is shown.

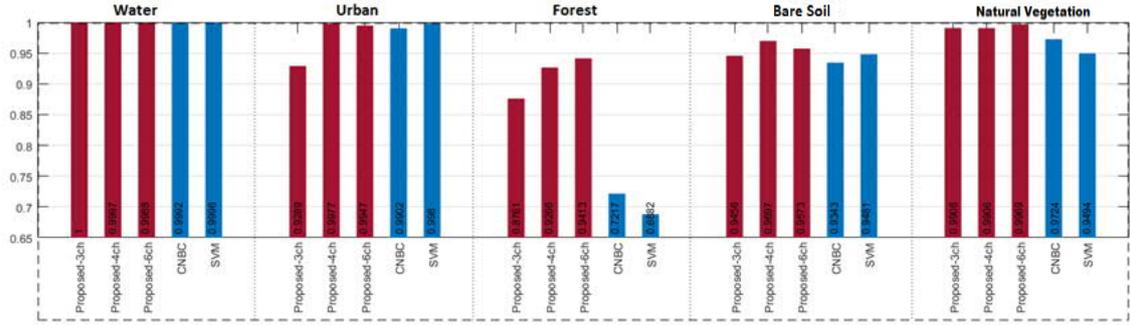

Figure 8. Classification accuracies of the proposed and competing methods for *SFBay_L*. $21 \times 21$ window size is used for the proposed approach and all EM features presented in Table 5 are used for the competing methods.

Table 8. Confusion Matrix is given with producer's and user's accuracies for each class and overall accuracy (OA) for *SFBay_L* obtained by the proposed approach with the 4-channel input and window size of 21.

|  |  | Predicted ||||| | Prod. Acc. |
|---|---|---|---|---|---|---|---|---|
|  |  | Water | Urban | Forest | Bare Soil | Nat. Vegetation | Total |  |
| True | Water | **78621** | 0 | 0 | 0 | 27 | 78648 | 99.97% |
|  | Urban | 0 | **17940** | 38 | 4 | 0 | 17982 | 99.77% |
|  | Forest | 0 | 313 | **4202** | 0 | 20 | 4535 | 92.66% |
|  | Bare Soil | 78 | 71 | 68 | **6956** | 0 | 7173 | 96.97% |
|  | Nat. Vegatation | 0 | 0 | 128 | 0 | **13531** | 13659 | 99.06% |
|  | Total | 78699 | 18324 | 4436 | 6960 | 13578 | **121997** |  |
|  | User Accuracy | 99.90% | 97.90% | 94.72% | 99.94% | 99.65% | **OA:** 99.39% ||

*Performance Evaluations over SFBay_C*

For this study site, the window size is varied from $5 \times 5$ to $19 \times 19$ pixels, and as before, three different setups are used with 3 to 6 channels. Considering the results presented in Table 9, the proposed approach using the diagonal elements of $[T]$ with $19 \times 19$ window size achieves a significantly better accuracy (the gap is greater than 10%) compared to the best results of the competing method in (Uhlmann and Kiranyaz 2014) with a 46-D EM feature vector. Furthermore, this advantage is also seen when the competing method uses a 187-D composite feature vector with many color and texture features: >3% higher accuracy is achieved by the proposed approach with the only 4-D feature vector. This is a significant accomplishment considering the computational burden involved in producing a composite feature vector in such a high dimension.

Table 9. Classification accuracies of the proposed approach for *SFBay_C* with different window sizes and channels.

| SFBay (C Band) | 3-channels | 4-channels | 6-channels |
|---|---|---|---|
| **Window Size** | $T_{11}, T_{22}, T_{33}$ | $T_{11}, T_{22}, T_{33}, span$ | $T_{11}, T_{22}, T_{33}, C_{11}, C_{22}, C_{33}$ |
| **5x5** | 0.8417 | 0.8415 | 0.8324 |
| **7x7** | 0.883 | 0.8794 | 0.8687 |
| **9x9** | 0.9119 | 0.9136 | 0.9017 |
| **11x11** | 0.931 | 0.9292 | 0.922 |
| **13x13** | 0.938 | 0.9389 | 0.9312 |
| **15x15** | 0.944 | 0.9439 | 0.9395 |
| **17x17** | 0.9496 | 0.9466 | 0.9417 |
| **19x19** | **0.9496** | **0.9532** | **0.9452** |

The confusion matrix produced by the proposed approach for *SFBay_C* is given in Table 10, and a detailed comparison is provided by considering the classification accuracies per class in Figure 9. Similar observations can be made on the results, i.e., the proposed approach achieves certain performance improvements in all classes, and the gap can be significant on some, e.g. high/low urban classes. Note that the classification performance for water class is not improved for *SFBay_L* and *SFBay_C*, as illustrated in Figure 8 and Figure 9. This might be expected since the achieved accuracy for water was already approaching one in the competing methods; with limited scope for further improvements in practice, as observed in the provided confusion matrices.

For particular classes, the method in (Uhlmann and Kiranyaz 2014) is able to achieve a certain level of classification accuracy (i.e., > 80%), but only when all the extracted features (EM+Color+Texture) are used together in a 187-D composite feature vector. As before, such a high computational complexity caused by multiple feature extraction operations and the usage of a large ensemble of classifiers can be a disadvantage for certain real-time applications.

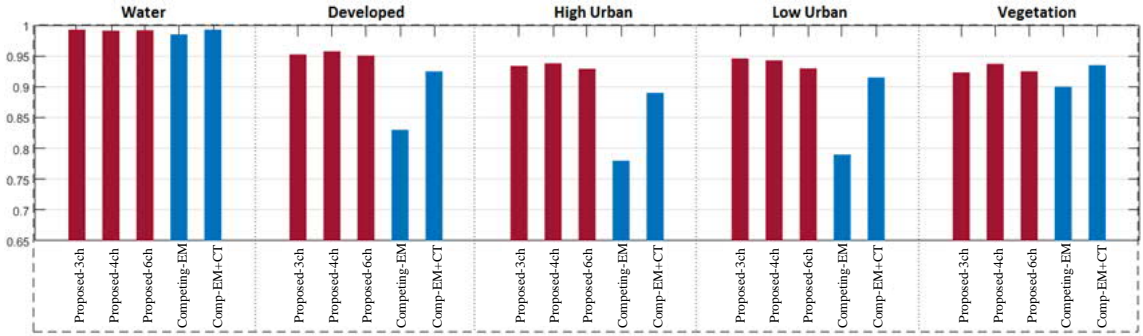

Figure 9. Classification accuracies of the proposed and competing methods for *SFBay_C*. 19 × 19 window size is used in the proposed approach. The competing method uses combinations of 46-D EM features given in Table 5 and a 187-D composite feature vector with EM, color, and texture features (EM + CT).

Table 10. Confusion Matrix is given with producer's and user's accuracies for each class and overall accuracy (OA) for *SFBay_C* obtained by the proposed approach with the 4-channel input and window size of 19.

|  |  | Predicted |  |  |  |  |  | Producer Acc. |
|---|---|---|---|---|---|---|---|---|
|  |  | Water | Developed | High-Den. Urban | Low-Den. Urban | Vegetation | Total |  |
| True | Water | **49545** | 1 | 108 | 59 | 287 | 50000 | 99.09% |
|  | Developed | 0 | **47873** | 100 | 523 | 1504 | 50000 | 95.75% |
|  | High-Den. Urban | 0 | 412 | **46901** | 2228 | 459 | 50000 | 93.80% |
|  | Low-Den. Urban | 219 | 124 | 1578 | **47131** | 948 | 50000 | 94.26% |
|  | Vegetation | 183 | 941 | 437 | 1579 | **46860** | 50000 | 93.72% |
|  | Total | 49947 | 49351 | 49351 | 49124 | 50058 | **250000** |  |
| User Accuracy |  | 99.20% | 97.01% | 95.04% | 95.94% | 93.61% | **OA:** 95.32% |  |

For visual evaluation, the training samples and the final segmentation masks for *SFBay_C* are shown in Figure 10. Similar arguments as in *SFBay_L* can be made on the effect of different window sizes: a coarse resolution on the segmentation mask occurs on large windows (e.g., *N*=17 and *N*=19), although the best classification accuracy is obtained by *N*=19.

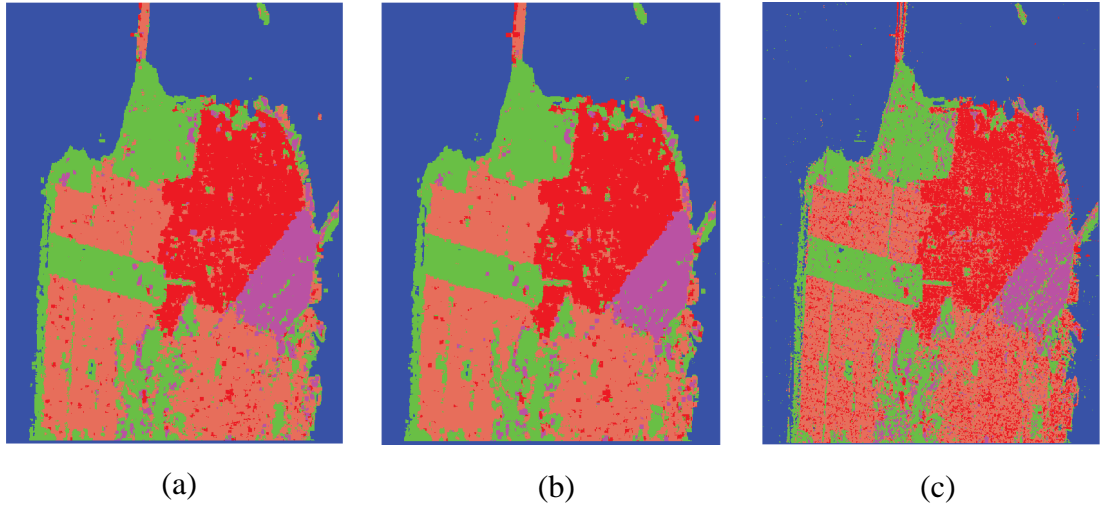

(a) (b) (c)

Figure 10. Final segmentation masks by the proposed approach for *SFBay_C*. The segmentation masks with 3- and 4-channel input using 17 × 17 and 19 × 19 pixels windows are shown in (a) and (b), respectively. In (c), the segmentation mask with 3-channel input using 7 × 7 pixels window is shown.

*Performance Evaluations over Flevo_L*

The classification task becomes more challenging compared to SFBay cases due to the 15 classes of natural land cover types for this study site. This provides an opportunity to test the performance of the proposed approach with respect to the increasing number of training samples. For this purpose, the number of pixels per class has been varied between 120 and 480 with an addition of an adaptive mode (*Adpt*) where 2% of each class is randomly selected for training. As presented in Table 11, the most promising accuracies for *Flevo_L* are obtained with the window sizes 7 × 7 or 9 × 9 pixels. The achieved best accuracy is 92.49%, obtained by using 9 × 9 pixel windows and 6-channel input (diagonal elements of [*T*] and [*C*]) using 480 training samples from each class. Confusion Matrix for *Flevo_L* produced using the 6-channel input and window size of 7 is given in Table 12.

Even though the number of per class samples is varied, 480 samples per class still correspond to only ~3.4% of the entire ground truth. This may be traded-off against having ~2% of data in training and 92.33% accuracy, which is an insignificant accuracy loss. This is similar to the train size used in (Uhlmann and Kiranyaz 2014), where the best accuracy is <83% with the feature set, FS1, in 12 dimensions (using elements from [T] and [C]). Note that the best average accuracy achieved by the study in (Uhlmann and Kiranyaz 2014) is approximately 91.5%, again using a composite feature set including 46-D EM + 60-D color and 81-D texture features with a total of 187-D feature vector and a large ensemble of RF classifiers. The proposed approach surpasses this level using only 6-D EM features and a compact CNN.

Table 11. Classification accuracies of the proposed approach for *Flevo_L* with different window sizes and channels.

| Flev.-L Window Size | 3-channels $T_{11}, T_{22}, T_{33}$ | 4-channels $T_{11}, T_{22}, T_{33}, span$ | 6-channels $T_{11}, T_{22}, T_{33}, C_{11}, C_{22}, C_{33}$ | #sample per class |
|---|---|---|---|---|
| **5x5** | 0.8633 | 0.8594 | 0.9014 | 120 |
| **7x7** | 0.8806 | 0.891 | 0.91 | 120 |
| **7x7** | 0.8946 | 0.8955 | 0.9191 | 240 |
| **7x7** | 0.8994 | 0.9006 | 0.9186 | 360 |
| **7x7** | 0.9022 | 0.9036 | 0.9227 | 480 |
| **7x7** | 0.904 | 0.9018 | **0.9233** | *Adpt.* |
| **9x9** | 0.8968 | 0.8837 | 0.9073 | 120 |
| **9x9** | 0.8968 | 0.8842 | 0.9133 | 240 |
| **9x9** | 0.8971 | 0.9014 | 0.9235 | 360 |
| **9x9** | 0.9 | 0.9043 | **0.9249** | 480 |
| **9x9** | 0.9033 | 0.903 | 0.9203 | *Adpt.* |
| **11x11** | 0.8765 | 0.87 | 0.8896 | 120 |
| **11x11** | 0.8915 | 0.8923 | 0.9105 | *Adpt.* |
| **21x21** | 0.7964 | 0.7992 | 0.8133 | 120 |
| **21x21** | 0.835 | 0.8368 | 0.8496 | *Adpt.* |
| **31x31** | 0.764 | 0.7245 | 0.7795 | 120 |
| **31x31** | 0.8465 | 0.8167 | 0.8524 | *Adpt.* |

Table 12. Confusion Matrix is given with producer's and user's accuracies for each class and overall accuracy (OA) for *Flevo_L* obtained by the proposed approach with the 6-channel input and window size of 7. The class order is as follows: water, forest, lucerne, grass, rapeseed, beet, potatoes, peas, stem beans, bare soil, wheat A, wheat B, wheat C, barley, and building.

| | | | | | | | Predicted | | | | | | | | | Tot. | Prod. Acc. |
|---|---|---|---|---|---|---|---|---|---|---|---|---|---|---|---|---|---|
| True | **28064** | 19 | 0 | 57 | 0 | 1 | 45 | 18 | 19 | 426 | 1 | 9 | 2 | 4 | 7 | 28672 | 97.88% |
| | 75 | **14565** | 0 | 27 | 0 | 117 | 723 | 0 | 13 | 0 | 4 | 0 | 1 | 0 | 19 | 15544 | 93.70% |
| | 272 | 0 | **9790** | 745 | 7 | 0 | 1 | 0 | 3 | 3 | 0 | 88 | 57 | 11 | 0 | 10977 | 89.19% |
| | 0 | 0 | 320 | **8896** | 0 | 173 | 171 | 1 | 207 | 3 | 0 | 9 | 133 | 238 | 50 | 10201 | 87.21% |
| | 77 | 0 | 3 | 80 | **19147** | 232 | 3 | 167 | 54 | 58 | 1208 | 248 | 144 | 1 | 0 | 21422 | 89.38% |
| | 4 | 54 | 4 | 272 | 460 | **12740** | 535 | 89 | 235 | 15 | 9 | 0 | 13 | 65 | 64 | 14559 | 87.51% |
| | 79 | 786 | 26 | 218 | 87 | 733 | **18376** | 175 | 217 | 12 | 48 | 1 | 128 | 15 | 23 | 20924 | 87.82% |
| | 0 | 8 | 0 | 16 | 57 | 358 | 171 | **9221** | 17 | 1 | 340 | 0 | 0 | 0 | 0 | 10189 | 90.50% |
| | 0 | 100 | 1 | 222 | 3 | 100 | 51 | 1 | **7780** | 8 | 16 | 2 | 7 | 0 | 12 | 8303 | 93.70% |
| | 349 | 0 | 0 | 37 | 0 | 0 | 0 | 0 | 6 | **5765** | 0 | 0 | 2 | 32 | 0 | 6191 | 93.12% |
| | 0 | 0 | 0 | 17 | 358 | 37 | 0 | 93 | 17 | 0 | **16220** | 372 | 178 | 0 | 0 | 17292 | 93.80% |
| | 0 | 0 | 0 | 0 | 639 | 5 | 0 | 0 | 0 | 0 | 409 | **9266** | 100 | 0 | 0 | 10419 | 88.93% |
| | 0 | 0 | 37 | 9 | 109 | 10 | 0 | 0 | 6 | 0 | 214 | 123 | **21077** | 0 | 0 | 21585 | 97.65% |
| | 73 | 0 | 13 | 172 | 0 | 0 | 0 | 0 | 0 | 4 | 0 | 0 | 0 | **6962** | 0 | 7224 | 96.37% |
| | 0 | 2 | 0 | 8 | 0 | 0 | 0 | 0 | 0 | 0 | 0 | 0 | 0 | 0 | **511** | 521 | 98.08% |
| Tot. | 28993 | 15534 | 10194 | 10776 | 20867 | 14506 | 20076 | 9765 | 8574 | 6295 | 18469 | 10118 | 21842 | 7328 | 686 | 204023 | |
| User. Acc. | 96.80% | 93.76% | 96.04% | 82.55% | 91.76% | 87.83% | 91.53% | 94.43% | 90.74% | 91.58% | 87.82% | 91.58% | 96.50% | 95.01% | 74.49% | **OA: 92.33%** | |

The visual inspection over the segmentation masks continues revealing the performance improvements. Consider the best results of both methods using overlaid regions over the corresponding ground truth, as illustrated in Figure 11 and Figure 12. The proposed approach can generate significantly better segmentation masks with less or no noise especially for *water, rapeseed,* and *barley* classes, even though the competing method uses a significantly higher amount of information with a 187-D composite feature vector.

Previous discussions contained an analysis of the role of $N$ in terms of the classification performance, revealing that better accuracies were achieved by using large $N$ values which can be traded-off against having finer details in the segmentation masks. However, for this study site, the best quantitative results in terms of accuracy are obtained by using small $N$ values. Therefore, the optimal window size $N$ is highly dependent on the SAR data; for instance, *Flevo_L* is more heterogeneous compared to others in terms of number of class types.

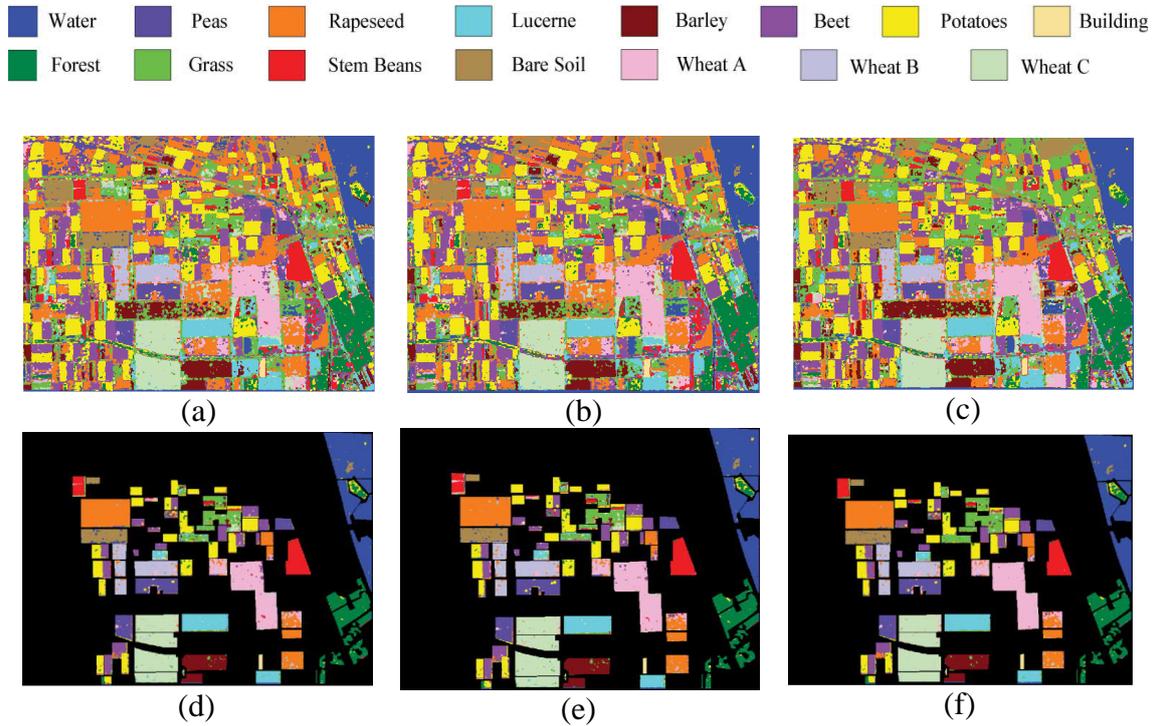

Figure 11. For *Flevo_L*, segmentation masks with 7x7 window size and of 3., 4., and 6. channels are shown in (a), (b), and (c), respectively. Their corresponding overlaid regions on the ground truth are shown in (d), (e), and (f), respectively.

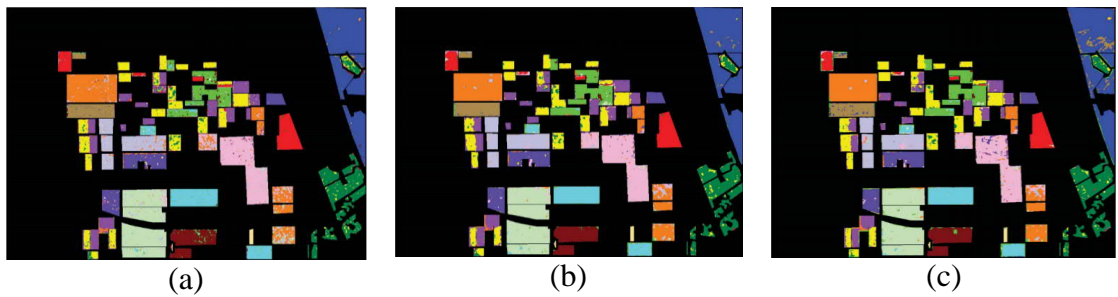

Figure 12. Segmentation masks over ground truth of the competing method in (Uhlmann and Kiranyaz 2014) using only EM features in (a), EM and Color in (b), and EM and Color+Texture in (c) for *Flevo_L*.

*Performance Evaluations over Flevo_C*

Classification accuracies of the proposed approach for this benchmark site with different window sizes and channels are presented in Table 13. The best accuracy achieved is 96.35% using 15 × 15 pixels window and 3-channel input, and its corresponding confusion matrix is given in Table 14. This is higher than the best classification accuracies in (Uhlmann and Kiranyaz 2014), which are 91.19% with 46-D EM features and 95.68% with 187-D composite (EM and Color+Texture) features for this study site.

Given the best results of both methods using overlaid regions over the corresponding ground truth, as illustrated in Figure 13 and Figure 14, the conclusions that were drawn for *SFBay_C* are also valid regarding the superiority of the proposed approach on classification performance and computational complexity. For example, the obtained mask using 46-D EM features in the competing method is not discriminative at all for *cropland*, which is misclassified as *forest* and *urban*. Moreover, when color and texture features are combined with EM features, the proposed model still leads in discriminative power, even though they use 187-D features.

Table 13. Classification accuracies of the proposed approach for *Flevo_C* with different window sizes and channels.

| Flev.-C Window Size | 3-channels $T_{11}, T_{22}, T_{33}$ | 4-channels $T_{11}, T_{22}, T_{33}, span$ | 6-channels $T_{11}, T_{22}, T_{33}, C_{11}, C_{22}, C_{33}$ |
|---|---|---|---|
| 5x5 | 0.9078 | 0.9034 | 0.9115 |
| 7x7 | 0.9357 | 0.9353 | 0.9367 |
| 9x9 | 0.9543 | 0.9539 | 0.9552 |
| 11x11 | 0.9575 | 0.9571 | 0.9537 |
| 13x13 | 0.963 | 0.9596 | 0.9561 |
| 15x15 | **0.9635** | **0.9631** | **0.9614** |

Table 14. Confusion Matrix is given with producer's and user's accuracies for each class and overall accuracy (OA) for *Flevo_C* obtained by the proposed approach with the 3-channel input and window size of 15.

| | | Predicted | | | | | Producer Acc. |
|---|---|---|---|---|---|---|---|
| | | Water | Urban | Forest | Cropland | Total | |
| True | Water | **49616** | 287 | 27 | 70 | 50000 | 99.23% |
| | Urban | 13 | **48489** | 701 | 797 | 50000 | 96.98% |
| | Forest | 44 | 1539 | **46664** | 1753 | 50000 | 93.33% |
| | Cropland | 89 | 632 | 1346 | **47933** | 50000 | 95.87% |
| | Total | 49762 | 50947 | 48738 | 50553 | **200000** | |
| User Accuracy | | 99.71% | 95.18% | 95.74% | 94.82% | **OA:** 96.35% | |

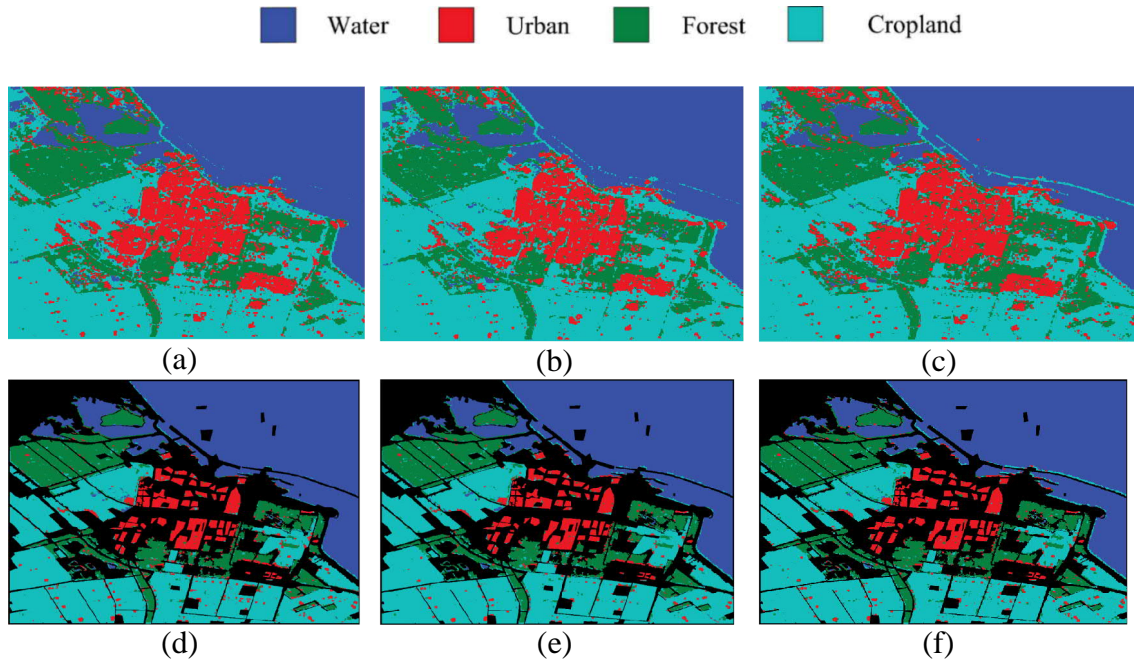

Figure 13. Segmentation masks with 15 × 15 window size and of 3., 4., and 6. channels are shown in (a), (b), and (c), respectively, for *Flevo_C*. Their corresponding overlaid regions on the ground truth are shown in (d), (e), and (f), respectively.

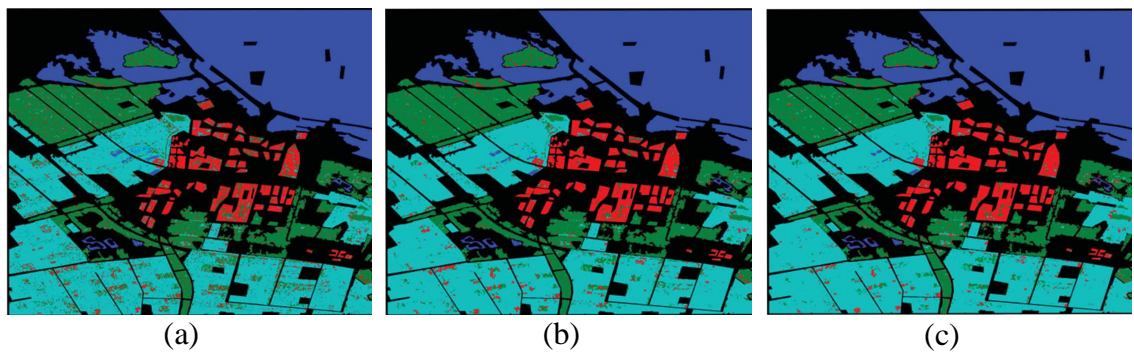

Figure 14. Segmentation masks of the competing method in (Uhlmann and Kiranyaz 2014) using only EM features in (a), EM and Color in (b), and EM and Color+Texture in (c) for *Flevo_C*.

*Generalization Capability Analysis: Cross-site Validation*

Since the proposed approach is based on a compact CNN configuration, it would be expected to have superior generalization capability. For example, because the size of the training dataset is significantly smaller compared to the test dataset size, the over-fitting would be expected for the deep CNNs causing poor classification performances on the test set. However, observations of the experiments have shown satisfactory classification performances of the proposed approach over four different study sites, despite the limited size of the training set. Additionally, the generalization capability of the proposed approach is further evaluated. First, the training procedure is performed separately for two different networks using training samples of *SFBay_C* and *Flevo_C*, and then, the trained networks are evaluated over test samples of different study sites, *SFBay_L* and *Flevo_L*, without further training and fine-tuning.

  The confusion matrices computed over *SFBay_L* and *Flevo_L* are presented in Table 15 and Table 16, respectively. To perform such a cross-site validation, only the common class types between training and validation sites are considered in the performance analysis. Accordingly, the pre-trained network over *SFBay_C* has 5-neurons (for water, developed, high-density urban, low-density urban, and vegetation classes) in the output layer. To make inference possible over *SFBay_L* using the pre-trained network over *SFBay_C*, the following classes are considered as a single class of urban in *SFBay_L*: developed, high-density urban, and low-density urban classes. Similarly, the confusion matrices computed over *Flevo_L* are given in Table 16. Unlike *Flevo_C*, *Flevo_L* consists of many different crop types, and therefore, four crop types are merged into a single cropland class for the analysis: rapeseed, wheat B, wheat C, and beet. Other crop types are removed from the evaluation since they are not present in the *Flevo_C* site. The obtained accuracies indicate significant performance degradation in the cross-site validation. However, the decrement in the performance is expected in the cross-site validation since the training and validation study sites do not consist of perfectly overlapping regions, even though they correspond the similar areas. Moreover, the acquisition systems are also different for the training and validations sites: space-borne and air-borne, respectively. Nevertheless, in practical usage, separate training procedures would be performed for the classification of each PolSAR image. In fact, this study may be the first that uses cross-site validation to evaluate the generalization capability of CNN-based SAR classification approaches.

Table 15. The computed confusion matrix is given with producer's and user's accuracies for each class and overall accuracy (OA) for *SFBay_L* obtained by the proposed approach that is trained over *SFBay_C* with the 4-channel input and window size of 21. The obtained accuracy is 76.01%.

|  |  | Predicted | | | | Producer Accuracy |
|---|---|---|---|---|---|---|
|  |  | **Water** | **Urban** | **Forest** | **Total** |  |
| True | **Water** | **56137** | 22511 | 0 | 78648 | 71.38% |
|  | **Urban** | 0 | **17982** | 0 | 17982 | 100% |
|  | **Forest** | 0 | 1754 | **2781** | 4535 | 61.32% |
|  | **Total** | 56137 | 42247 | 2781 | **101165** |  |
| **User Acc.** |  | 100% | 42.56% | 100% | **OA: 76.01%** | |

Table 16. The computed confusion matrices are given with producer's and user's accuracies for each class and overall accuracy (OA) for *Flevo_L* obtained by the proposed approach that is trained over *Flevo_C* with the 6-channel input and window size of 7. Different crop types of *Flevo_L* are combined into a single class: cropland. In (a), cropland consists of rapeseed and wheat B, in (b), it consists of rapeseed, beet, wheat B, and wheat C. The obtained accuracies are 88.11% and 77.35% in (a) and (b), respectively.

(a)

|  |  | Predicted | | | | | Prod. Acc. |
|---|---|---|---|---|---|---|---|
|  |  | **Water** | **Forest** | **Building** | **Cropland** | **Total** |  |
| True | **Water** | **21590** | 3 | 894 | 6185 | 28672 | 75.30% |
|  | **Forest** | 0 | **15419** | 103 | 22 | 15544 | 99.20% |
|  | **Building** | 0 | 0 | **521** | 0 | 521 | 100% |
|  | **Cropland** | 705 | 663 | 528 | **29945** | 31841 | 94.05% |
|  | **Total** | 22295 | 16085 | 2046 | 36152 | **76578** |  |
| **User Acc.** |  | 96.84% | 95.86% | 25.46% | 82.83% | **OA: 88.11%** | |

(b)

|  |  | Predicted | | | | | Prod. Acc. |
|---|---|---|---|---|---|---|---|
|  |  | **Water** | **Forest** | **Building** | **Cropland** | **Total** |  |
| True | **Water** | **21590** | 3 | 894 | 6185 | 28672 | 75.30% |
|  | **Forest** | 0 | **15419** | 103 | 22 | 15544 | 99.20% |
|  | **Building** | 0 | 0 | **521** | 0 | 521 | 100% |
|  | **Cropland** | 9623 | 7107 | 1594 | **49661** | 67985 | 73.05% |
|  | **Total** | 31213 | 22529 | 3112 | 55868 | **112722** |  |
| **User Acc.** |  | 69.17% | 68.44% | 16.74% | 88.89% | **OA: 77.35%** | |

*Sensitivity Analysis on Hyper-parameters*

In the proposed adaptive CNNs, a compact 4-layer network configuration is used as presented. In this section, the hyper-parameter sensitivity of the network is analyzed by varying them significantly. For this purpose, *m* is defined as the multiplier for the number of neurons in hidden layers and *n* as the multiplier for the number of hidden CNN layers, respectively. For example, *m*=*n*=1 corresponds to the default setup and if *m*=4 and *n*=2, then, the network would have two hidden CNN layers with 4x20=80 neurons each and 4x10=40 neurons for the hidden MLP layer (i.e., [*In-80-80-40-Out*]).

The classification accuracies are given in Table 17. Accordingly, the optimal number of channels and window size *N* are fixed for each PolSAR image, while only the network hyper-parameters are varied. It is observed in Table 17 that as the network configuration becomes deeper and more complex, accuracies tend to decrease. This is an expected outcome due to the well-known "Over-Fitting" phenomenon. For *m*=8, *n*=1 and *m*=8, *n*=2 models, no convergence has occurred at all with 5 BP runs each with 1000 iterations.

Table 17. Classification accuracies of the eight network configurations are given using different number of hidden neurons (multiplier *m*) and hidden CNN layers (multiplier *n*), where *m*=*n*=1 stands for initial (proposed) configuration. Entries with "---" means no convergence has occurred during BP training.

|  | *SFBay_L* | *SFBay_C* | *Flevo_L* | *Flevo_C* |
|---|---|---|---|---|
| m=1, n=1 | **0.9939** | 0.9532 | 0.9233 | 0.9635 |
| m=1, n=2 | 0.9753 | 0.9371 | 0.8967 | 0.9503 |
| m=2, n=1 | 0.9931 | **0.956** | **0.9305** | **0.9641** |
| m=2, n=2 | 0.9615 | 0.9318 | 0.8611 | 0.9482 |
| m=4, n=1 | 0.9862 | 0.9498 | 0.8944 | 0.9687 |
| m=4, n=2 | 0.8245 | 0.936 | 0.7898 | 0.9463 |
| m=8, n=1 | 0.7366 | 0.4214 | --- | --- |
| m=8, n=2 | 0.6019 | --- | --- | --- |

Overall, the proposed approach is robust against certain variations in the number of hidden neurons. For example, in three study sites, the classification accuracies remain almost the same for m=1 and m=2. However, if the network increases in complexity (e.g., *m*>2) or depth (*n* >1), then, the generalization capability of the CNN tends to decrease as a natural consequence of the over-fitting.

**Conclusions and Future Work**

In this study, a novel approach is proposed for the classification of PolSAR data. The proposed system is based on compact and adaptive CNNs. In contrast to the deep counterparts requiring massive data sizes for training, they can successfully learn and generalize on limited training data and using smaller patches. The former property yields a crucial advantage in terms of usability with minimal human intervention and computational complexity that makes the proposed approach very suitable for real-time applications. The latter property is especially unique and important for the classification of fine spatial resolution SAR images with an improved segmentation resolution and accuracy, which occurs because the correlation of pixels disappears as the window size increases. Moreover, such deep networks require special hardware, unlike the proposed approach, which can conveniently be performed on ordinary computers. The competing method proposed in (Uhlmann and Kiranyaz 2014) requires the extraction of a large number of EM, texture, and color features (in 187-D) in advance to use in a large and complex ensemble of classifiers. In contrast, the experimental results demonstrate that using only 3 to 6 EM features, the proposed approach with a compact CNN can achieve a superior classification performance. One observation is that the window size, *N*, is the important hyper-parameter of the proposed approach; however, similar performance levels in a narrow margin can usually be achieved when *N* is set to any value from 7 to 19.

In this study, four benchmark PolSAR sites acquired at L and C bands are used; however, the proposed approach may be investigated over new bands. In order to further improve the performance, new ways are investigated to combine image processing features with EM channels within the proposed CNN model. It is important to note that only real-valued EM channels are used in this study, while the complete information of the surface is actually acquired using complex channels. These issues will be further explored in future work.